\documentclass[runningheads]{llncs}
\usepackage[T1]{fontenc}
\usepackage{amsmath}
\usepackage{graphicx,verbatim}
\usepackage{url}
\usepackage[pagebackref,breaklinks,colorlinks]{hyperref}
\usepackage[utf8]{inputenc}
\usepackage[small]{caption}
\usepackage{graphicx}
\usepackage{amsmath}
\usepackage{makecell}
\usepackage{booktabs}
\usepackage{array}
\usepackage[table]{xcolor}
\definecolor{azure}{rgb}{0.0, 0.5, 1.0}
\definecolor{alizarin}{rgb}{0.82, 0.1, 0.26}

\usepackage{amsthm}

\usepackage{booktabs}
\usepackage{algorithm}
\usepackage{algpseudocode}       % algorithmicx + \State, \If, \Comment
\usepackage[switch]{lineno}
\usepackage{enumitem}
\usepackage{fontawesome}
\usepackage[most]{tcolorbox}
\usepackage[table,xcdraw]{xcolor}
\usepackage{colortbl}
\usepackage{amssymb}
\usepackage{makecell}
\usepackage{subcaption}
\usepackage{rotating}

\definecolor{myblue}{HTML}{ACE5EE}
\definecolor{lapGreen}{HTML}{D5E8D4} % soft green
\definecolor{giBlue}{HTML}{DAE8FC}   % soft blue%

\newcommand{\sqimg}[2]{\includegraphics[width=#1,trim=60 40 60 40,clip]{#2}}

\newtcolorbox{HighlighterBox}[2][]{
    arc=3.8pt,
    left=12.0pt,
    right=12.0pt,
    bottom=4pt,
    top=4pt,
    colback=yellow!4,
    colframe=yellow!0,
    boxrule=0.8pt,
    colbacktitle=yellow!15,
    coltitle=black,
    title=\text{#2},
    #1,
    breakable,
    enhanced jigsaw
}
\newtcolorbox{highlighterbox}[1][]{
    arc=1.8pt,
    left=8.0pt,
    right=6.0pt,
    bottom=2pt,
    top=3pt,
    rounded corners,
    boxrule=0.8pt,
    colframe=teal!60,
    colback=teal!5,
    breakable,
    enhanced jigsaw
}

\begin{document}

\title{On the Robustness of Temporal Vision-Language Models for Surgical Endoscopy Videos}
\titlerunning{Temporal Robust Endoscopy CLIP}
\author{Darakshan Rashid$^{*1,2}$ \and
Raza Imam$^{*,1}$ \and
Ufaq Khan$^1$ \and
Muhammad Bilal$^3$ \and
Shazad Ashraf$^4$ \and
Dwarikanath Mahapatra$^5$ \and
Mohammad Yaqub$^1$ \and
Muhammad Haris Khan$^1$ \and
Imran Razzak$^1$ \and
Brejesh Lall$^2$ \and
Lena Maier-Hein$^{1,6}$ \and
Yutong Xie$^1$ 
}
\authorrunning{Darakshan et al.}
\institute{ Mohamed bin Zayed University of Artificial Intelligence (MBZUAI), UAE \\
\email{yutong.xie678@gmail.com} \and Indian Institute of Technology Delhi, India  \and Birmingham City University, UK \and  University Hospitals Birmingham, UK \and Khalifa University, UAE \and German Cancer Research Center (DKFZ), Germany}
\renewcommand{\thefootnote}{}%
\footnotetext{
Dataset and Code is available at: \href{https://github.com/DarakshanRashid/RobustEndoCLIP}{Github}
}

\footnotetext{
Corresponding Author: Yutong Xie 
%\Letter 
}%
% \email{bsz228540@iitd.ac.in} 
\footnotetext{
Equal Contribution * \hfill Accepted to MICCAI 2026
}%

\renewcommand{\thefootnote}{\arabic{footnote}}

\maketitle

\begin{abstract}
% Temporal vision-language models (TVLMs) offer a reusable interface for surgical video understanding via prompt-based clip classification.
% In endoscopy, however, deployment conditions are dominated by acquisition and pipeline artifacts: defocus, haze/smoke, motion blur, noise, and transmission distortions, which can break this assumed generality.
Temporal vision-language models (TVLMs) offer a reusable, prompt-based interface for surgical video understanding, yet, their robustness under clinically realistic acquisition artifacts in endoscopy remains insufficiently characterized. In practice, degradations such as defocus, haze, motion blur, noise, cautery smoke, and packet loss introduce structured distribution shifts which may compromise video–text alignment.
We study the robustness of temporal VLMs under such shifts caused by corruptions in clip frames.
We introduce \textbf{Endo-C6}, a compact corruption benchmark of six endoscopy-realistic perturbations evaluated at a fixed high severity, and apply it to public Gastrointestinal (GI) endoscopy and laparoscopic cholecystectomy videos.
Under a standardized prompt protocol, we benchmark 3 recent surgical TVLM baselines and analyze robustness in both mean and worst-case settings, spanning 294 dataset-level evaluations. Finally, we present \textbf{RobustEndoCLIP},  obtained by \textit{few-shot} parameter-efficient tuning with \textit{VeRA},
outperforming existing TVLM baselines.
Our findings show that off-the-shelf TVLMs can exhibit severe worst-case collapse under endoscopy-specific corruptions, whereas lightweight few-shot adaptation can substantially improve corrupted performance and robustness without changing the prompt-based interface. We expect Endo-C6 to support standardized robustness reporting and promote more reliable clinical vision-language systems. 

\keywords{Robustness \and Surgical video \and Endoscopy \and Vision-Language}
\end{abstract}

\section{Introduction}

% \textbf{Reusable TVLM interface.}
Temporal vision-language models (TVLMs) increasingly act as a \emph{software layer} for surgical video understanding \cite{yuan2025learning,yuan2024hecvl,yuan2024procedure,walimbe2025adaptation,honarmand2024vidlpro}: rather than training a bespoke classifier for each label set, a pretrained video-text representation can be queried with prompts and reused across tasks (phases, tools, events, findings).
This is attractive in surgical endoscopy{\footnote{In the remainder of the paper, ``endoscopy'' refers to surgical endoscopy.}, where annotation is costly, label ontologies drift across centers \cite{lavanchy2024challenges,reiter2023domain}, and much of the clinical value comes from detecting safety-critical events reliably \cite{wei2021intraoperative,mohamadipanah2023generating}.
{However, this reuse implicitly assumes video-text alignment remains stable under realistic operating conditions.}

\textbf{Endoscopy is intrinsically noisy:}
Endoscopy violates clean-benchmark assumptions
% Endoscopy rarely matches the ``clean'' imaging conditions assumed by standard benchmarks 
\cite{ali2021deep,ali2020objective,kim2025automated}. 
Signals are shaped by wet optics like lens fog, specular surfaces (glare), non-stationary illumination like overexposure, scope-cast shadows, and abrupt camera motion like motion blur and transient defocus, then further distorted by compression and sometimes live transmission like packet loss.
These artifacts are routine: models are relied upon during cautery smoke \cite{pan2022desmoke}, rapid repositioning, or low-light \cite{chen2024lightdiff} inspection, when defocus, lens fog \cite{lawrentschuk2010laparoscopic}, blur, and packet loss \cite{shor2023does} are most likely.
Thus, operational deployment requires robustness, meaning consistent behavior under label-preserving perturbations \cite{jaspers2024robustness,hendrycks2019benchmarking}.

\textbf{Why robustness is hard?}
Robustness benchmarks exist for natural vision \cite{hendrycks2019imagenetc} and are emerging for medical VLMs \cite{imam2025robustness}, but endoscopy needs robustness precisely under adverse conditions (smoke, blur/defocus, low light, compressed video) where rare failures can be high-stakes.
We study contrastively trained temporal video-text \emph{dual-encoders} for prompt-based
fixed-vocabulary clip classification, yet robustness is hard to compare: cross-dataset transfer confounds artifacts with shifts in procedure mix/labels, and video evaluations often underspecify clip length, stride, and temporal aggregation, which interact with corruptions and alter severity (Fig.~\ref{fig:main}(c)).
Thus, a compact, reproducible protocol that isolates artifact-driven failures while preserving temporal structure remains missing.

\textbf{Existing gap and direction:}
Prior surgical TVLMs transfer well under clean/curated settings \cite{yuan2025learning,yuan2024hecvl,yuan2024procedure}, and contrastive dual-encoder pretraining (e.g., CLIP) offers a general video-text alignment template \cite{radford2021learning}.
Robustness work shows medical VLMs can fail under perturbations and that parameter-efficient fine-tuning (PEFT) can help \cite{imam2025robustness,hu2022lora,kopiczko2023vera}, but existing endoscopy robustness studies target depth estimation on SCARED dataset and omit temporal prompt-based VLMs, GI-laparoscopy artifact diversity, and limited-supervision regimes \cite{wang2024benchmarking}.
We fill this gap by stress-testing TVLMs with endoscopy-realistic corruptions, reporting Mean/Worst-case performance, and comparing PEFT under a fixed label budget.  Endo-C6 evaluates prompt-based clip classification on CholecT50 surgical phases, Kvasir GI landmark/pathology classes, and TEMSET-24K surgical
actions. To this end, we formulate the following research questions:
% \textbf{Existing gap:}
% Prior surgical vision language  models demonstrate strong transfer under clean or curated settings \cite{yuan2025learning,yuan2024hecvl,yuan2024procedure}, and CLIP-style pretraining provides a template for video-text alignment \cite{radford2021learning}.
% Separately, robustness studies show that medical VLMs can degrade sharply under controlled perturbations and can benefit from parameter-efficient adaptation \cite{imam2025robustness,hu2022lora,kopiczko2023vera}.
% We note that the robustness of TVLMs on endoscopy is a missing piece that (i) stress-tests TVLMs under endoscopy-realistic artifacts, (ii) reports both mean and worst-case behavior, and (iii) evaluates PEFT choices under a fixed annotation budget. To address these gaps, we formulate the following research questions:
% Missing is a \emph{temporal endoscopy} robustness study that (i) stress-tests TVLMs under endoscopy-realistic artifacts, (ii) reports both mean and worst-case behavior, and (iii) evaluates PEFT choices under a fixed annotation budget. To address these gaps, we formulate the following research questions as:
\begin{enumerate}
    % \item[Q1.] Under Endo-C6 (our corruption benchmark), which corruption types dominate failure for zero-shot TVLMs?
    \item[Q1.] Which corruption types are most responsible for failure in zero-shot TVLMs?
    % \item[Q2.] Are corruption effects consistent across GI endoscopy vs.\ laparoscopic endoscopy, or domain-specific?
    \item[Q2.] Are corruption effects consistent across Gastrointestinal (GI) endoscopy vs.\ laparoscopic surgery, or domain-specific?
    \item[Q3.] Across models, does robustness track clean accuracy, or reflect distinct sensitivity (mean performance vs.\ worst-case collapse)?
    \item[Q4.] With a small label budget, does efficient few-shot tuning improve mean/worst-case robustness?
\end{enumerate}

% \textbf{Our blueprint:}
% We introduce Endo-C6, a compact corruption benchmark of six endoscopy-realistic perturbations evaluated at a fixed high severity, and apply it to 3 public GI endoscopy and laparoscopic cholecystectomy datasets.
% We benchmark SurgVLP \cite{yuan2025learning}, HecVLP \cite{yuan2024hecvl}, and PeskaVLP \cite{yuan2024procedure} under an identical prompt-based protocol, reporting mean and worst-case corrupted performance.
% Finally, we develop RobustEndoCLIP, an endoscopy-specialized temporal CLIP-style classifier obtained by few-shot parameter-efficient tuning with VeRA \cite{kopiczko2023vera} (with a matched LoRA comparison \cite{hu2022lora}) under fixed label budgets.

\textbf{Contributions:} We answer these questions with an in-depth analysis of existing TVLMs across endoscopic domains and highlight how to achieve robustness with our proposed approach. Specifically, we present our contributions:
% as follows:

\begin{enumerate}[leftmargin=1.6em]
    \item \textbf{Diverse Corruption Benchmark:} We define Endo-C6, a diverse \textit{endoscopy-video corruption benchmark}, comprising six endoscopy-realistic corruption types across GI endoscopy and laparoscopic surgery scenarios (See Fig.~\ref{fig:endoC6_taxonomy_stats},\ref{fig:endoc6_examples_frames}). 
    \item\textbf{Benchmarking Study on TVLMs:} We propose a reproducible prompt-based evaluation protocol for endoscopy video clip classification. Under this protocol, we benchmark representative surgical TVLM baselines: SurgVLP \cite{yuan2025learning}, HecVLP \cite{yuan2024hecvl}, and PeskaVLP \cite{yuan2024procedure}, on public endoscopy datasets under clean and corrupted conditions, \textit{analyzing} their fragility and robustness.
    \item \textbf{Robust Temporal VLM:} We develop RobustEndoCLIP, an endoscopy-specialized temporal CLIP-style classifier obtained by few-shot parameter-efficient tuning with VeRA adapters \cite{kopiczko2023vera}, and provide detailed analysis to conclude \textit{what drives robustness gains and where brittleness remains} in TVLMs.
\end{enumerate}

\begin{figure*}[t]
\centering

\begin{subfigure}[t]{0.35\textwidth}
  \vspace{0pt} % <-- forces top alignment
  \centering
  \includegraphics[width=\linewidth]{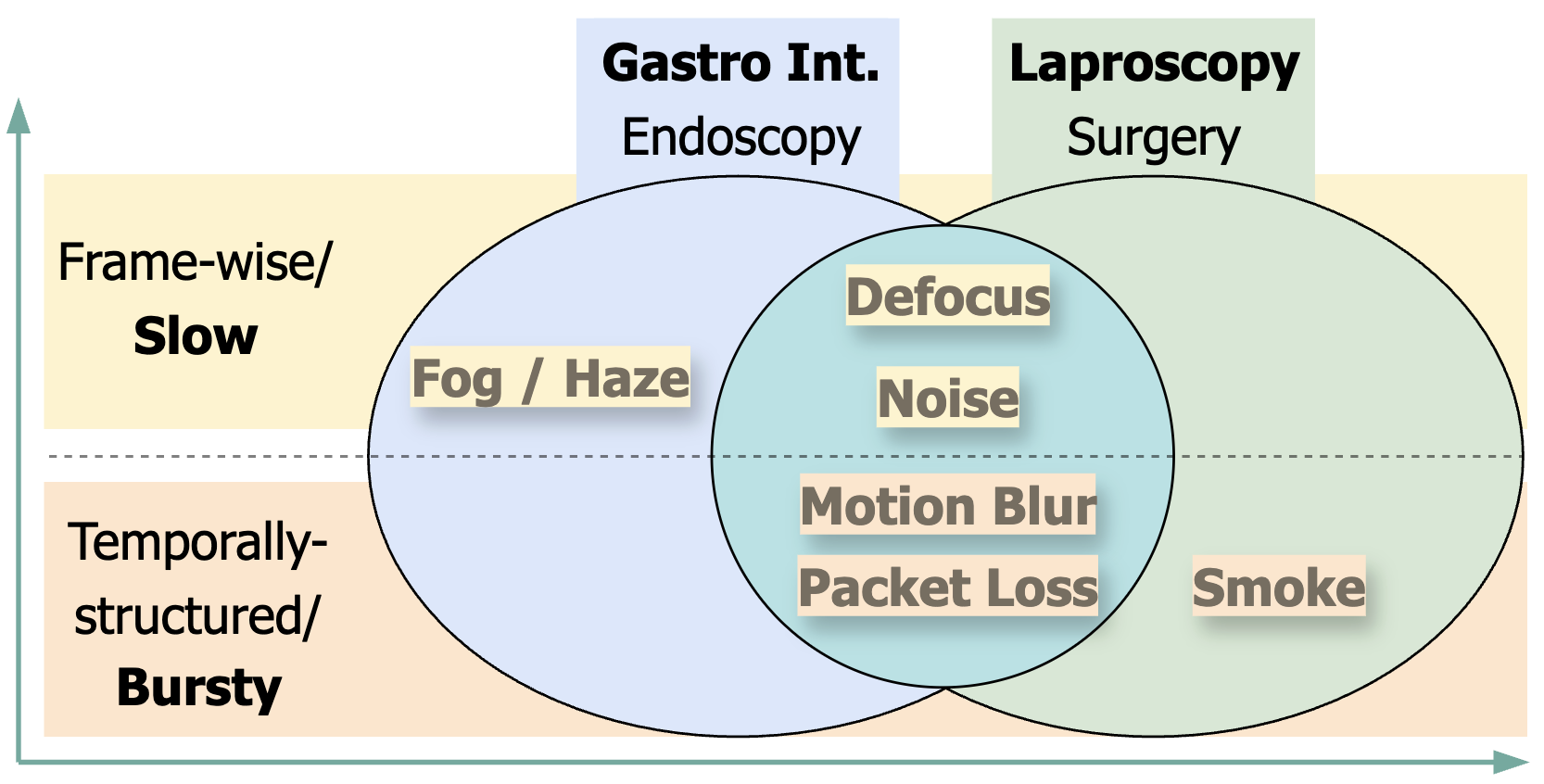}
  \caption{\small Our endoscopic taxonomy}
  \label{fig:endoC6_taxonomy}
\end{subfigure}
% \hfill
\begin{subfigure}[t]{0.40\textwidth}
  \vspace{0pt} % <-- forces top alignment
  \centering
  \small
  \setlength{\tabcolsep}{4pt}
  \renewcommand{\arraystretch}{1.12}

  % Use \linewidth, not 0.50\textwidth, since we're already inside a 0.49\textwidth subfigure
 
  \resizebox{\linewidth}{!}{%
  \begin{tabular}{c|c|ccc}
  \toprule
  \makecell{\textbf{Dataset} $\downarrow$} &
  \makecell{\textbf{Domain} $\downarrow$} &
  \makecell{\textbf{Train/Test}\\\textbf{\#Clips}} &
  \makecell{\textbf{Train/Test}\\\textbf{\#Frames}} &
  \makecell{\textbf{Frames}\\\textbf{per Clip}} \\
  \midrule
  \makecell{\colorbox{lapGreen}{\parbox{18mm}{\centering CholecT50\\~}}} &
  \makecell{\colorbox{lapGreen}{\parbox{18mm}{\centering Laparoscopic\\~}}} &
  \makecell{35/\\5} &
  \makecell{$\sim$70k/\\$\sim$10k} &
  \makecell{$\sim$2000} \\
  \midrule
  \makecell{\colorbox{giBlue}{\parbox{18mm}{\centering Kvasir\\~}}} &
  \makecell{\colorbox{giBlue}{\parbox{18mm}{\centering GI\\Endoscopy}}} &
  \makecell{8,000/\\1,200} &
  \makecell{8,000/\\1,200} &
  \makecell{1} \\
  \midrule
  \makecell{\colorbox{giBlue}{\parbox{18mm}{\centering TEMSET-24K\\~}}} &
  \makecell{\colorbox{giBlue}{\parbox{18mm}{\centering GI Colorectal\\TEMS}}} &
  \makecell{19,400/\\4,851} &
  \makecell{$\sim$582k/\\$\sim$146k} &
  \makecell{30} \\
  \bottomrule
  \end{tabular}%
  }
 
  \caption{\small Dataset statistics and splits}
  \label{fig:endoC6_stats}
\end{subfigure}

\caption{\small \textbf{Endo-C6 benchmark overview.} (a) Artifact taxonomy across endoscopic domains, organized by domain salience and temporal structure. (b) Dataset statistics across endoscopic domains covered in Endo-C6. Reported train-set counts follow the original dataset splits; unless otherwise stated, we train using only 16\% of the listed training data. All evaluations are conducted on the full test splits reported in (b).}
\label{fig:endoC6_taxonomy_stats}

\end{figure*}
% Requires in preamble:
% \usepackage{graphicx}
% \usepackage{rotating}

% Helper macro: square-ish crop (adjust trim if needed)
% \newcommand{\sqimg}[2]{%
% \includegraphics[width=#1,trim=60 40 60 40,clip]{#2}%
% }

\begin{figure*}[t]
\centering
\setlength{\tabcolsep}{2pt}
\renewcommand{\arraystretch}{1.0}

\resizebox{0.8\textwidth}{!}{
\begin{tabular}{c c c c c c c c}
% -------------------- Row 1: CholecT50 --------------------
\rotatebox{90}{\scriptsize{CholecT50}} &
\sqimg{0.145\textwidth}{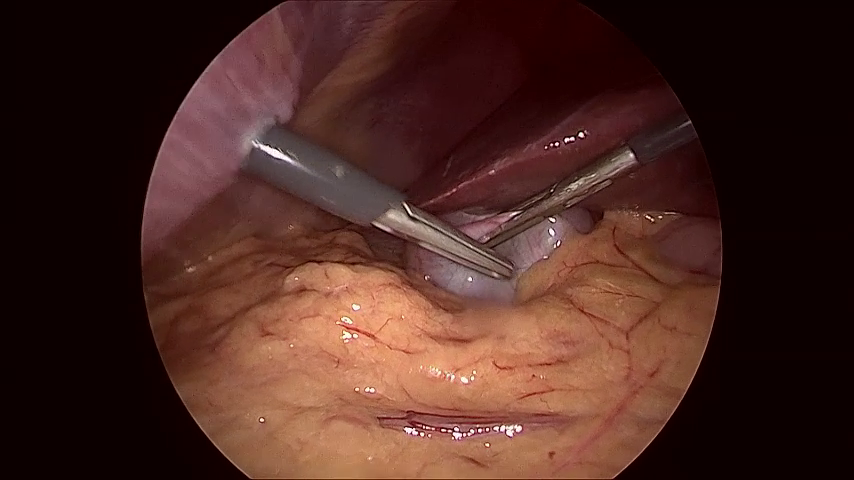} &
\sqimg{0.145\textwidth}{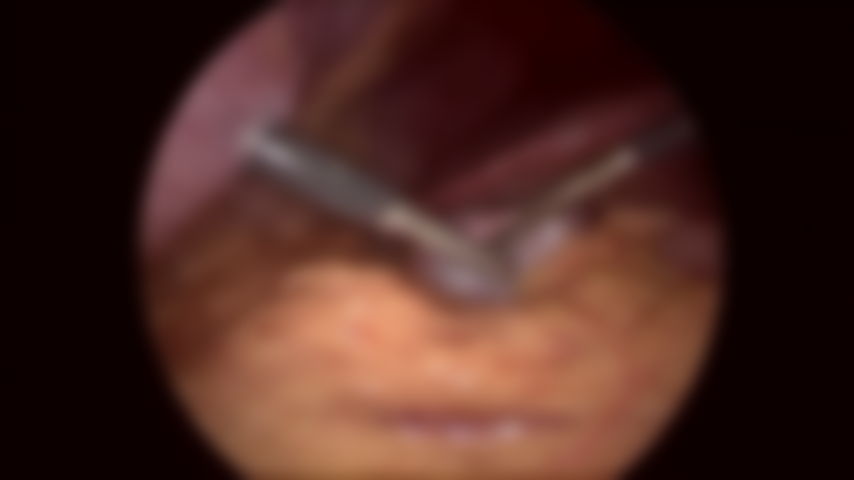} &
\sqimg{0.145\textwidth}{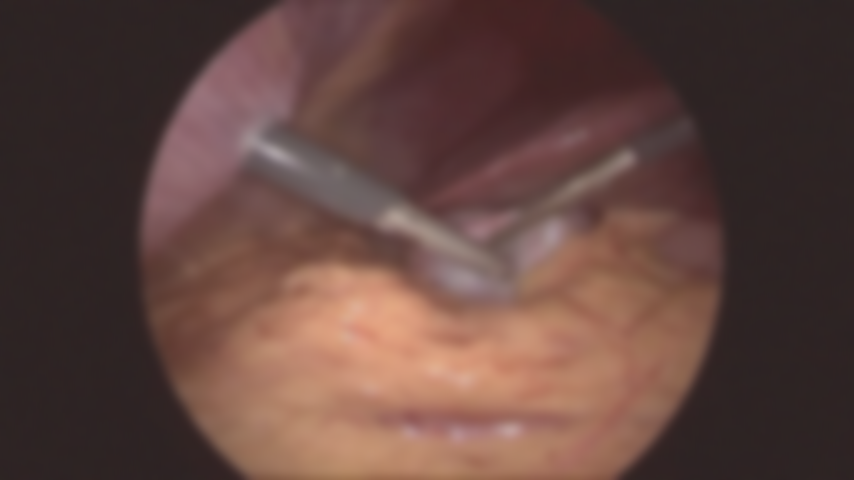} &
\sqimg{0.145\textwidth}{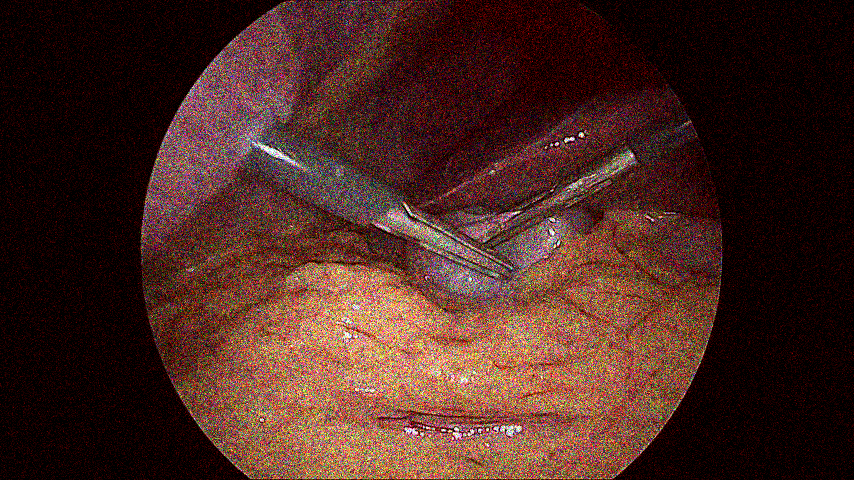} &
\sqimg{0.145\textwidth}{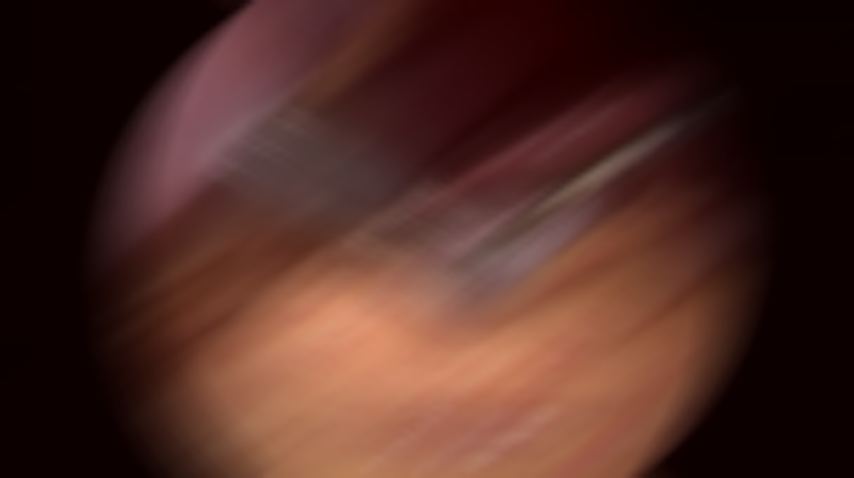} &
\sqimg{0.145\textwidth}{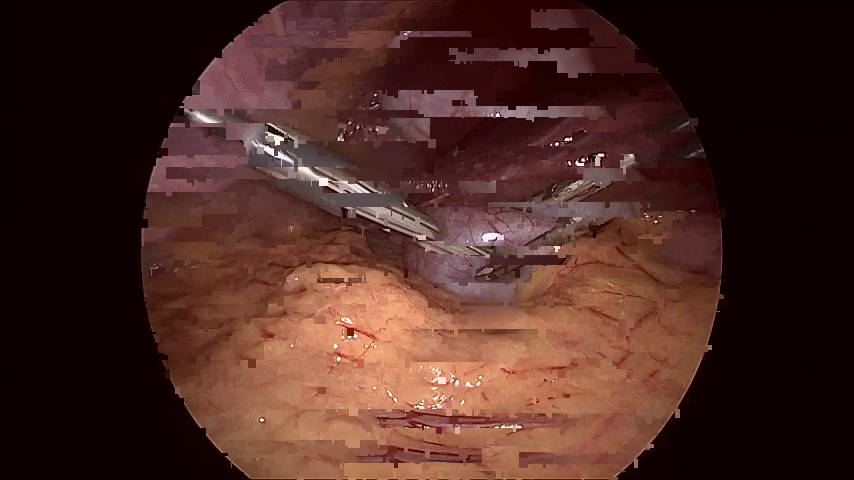} &
\sqimg{0.145\textwidth}{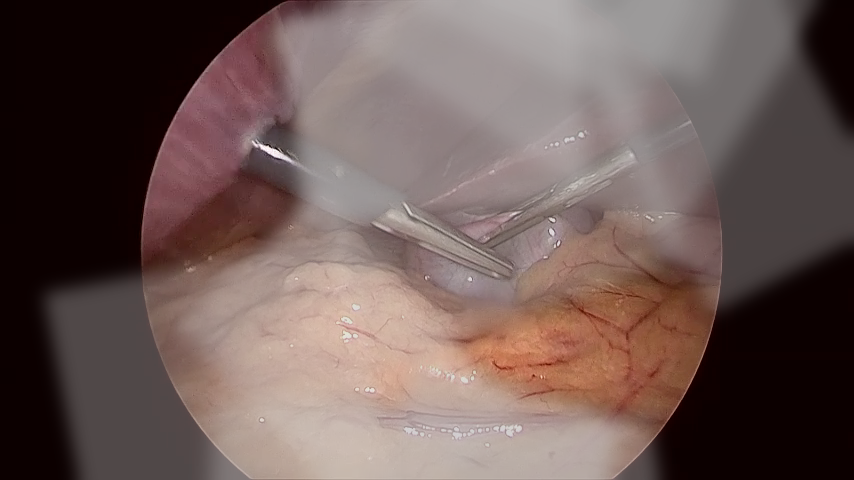} \\
% -------------------- Row 2: Kvasir --------------------
\rotatebox{90}{\scriptsize{Kvasir}} &
\sqimg{0.145\textwidth}{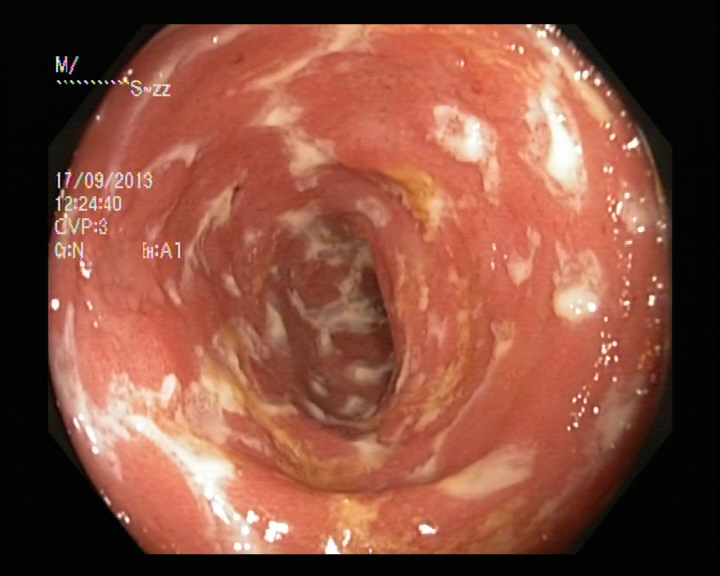} &
\sqimg{0.145\textwidth}{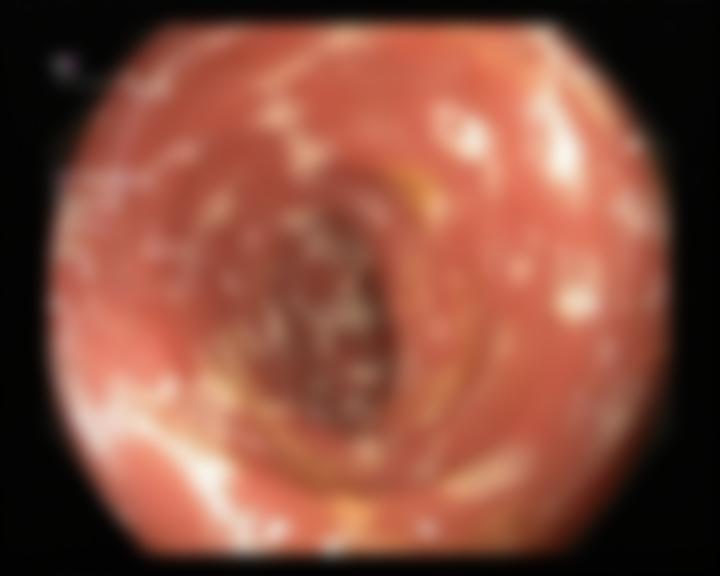} &
\sqimg{0.145\textwidth}{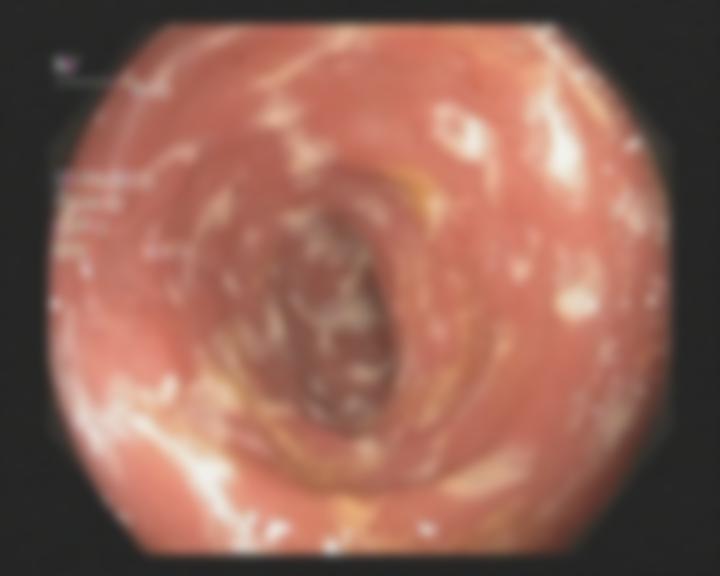} &
\sqimg{0.145\textwidth}{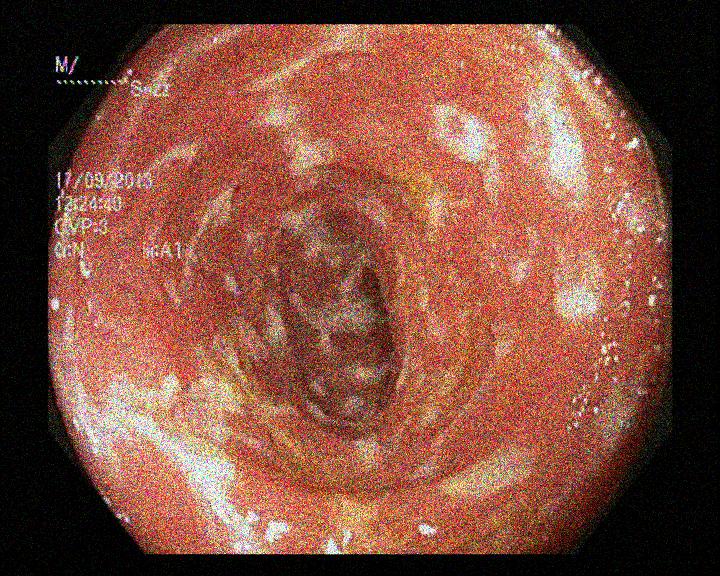} &
\sqimg{0.145\textwidth}{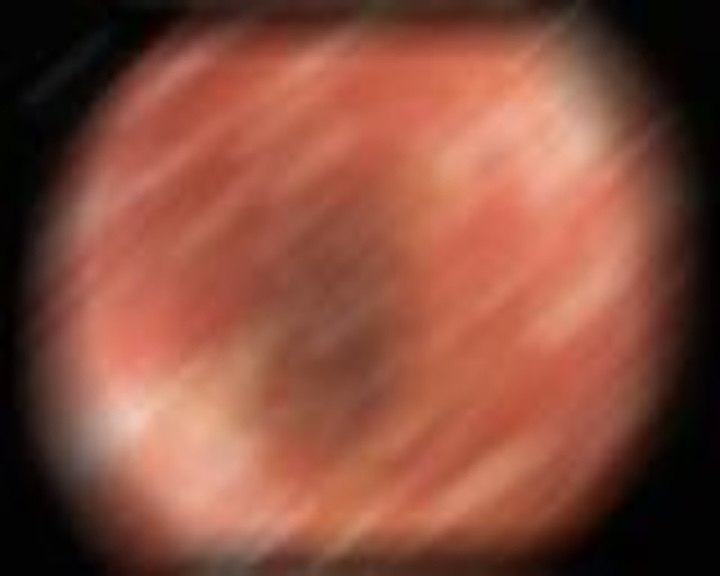} &
\sqimg{0.145\textwidth}{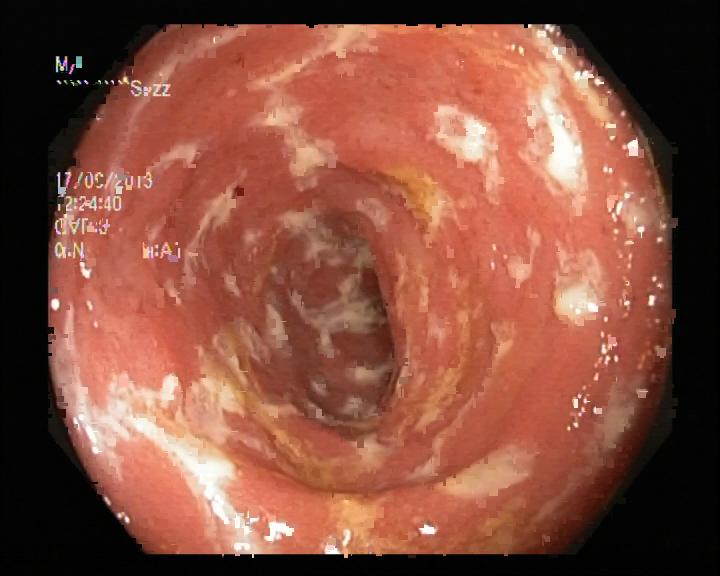} &
\sqimg{0.145\textwidth}{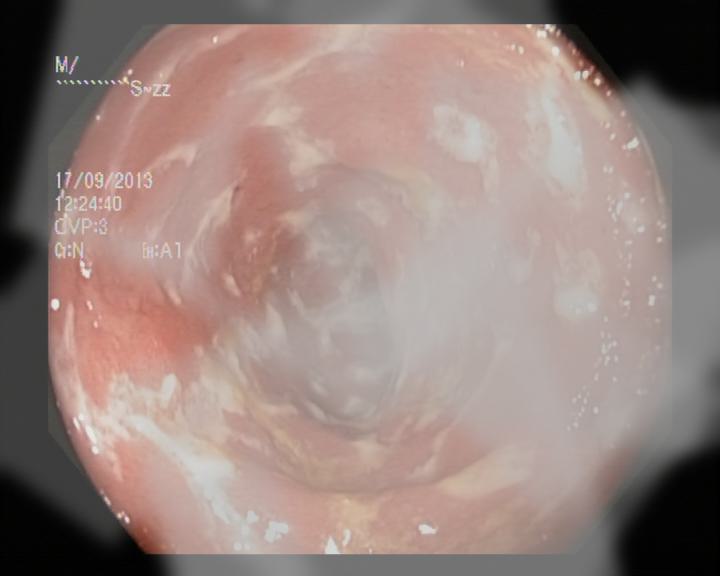} \\

% -------------------- Row 3: temset --------------------
\rotatebox{90}{\scriptsize{Temset}} &
\sqimg{0.145\textwidth}{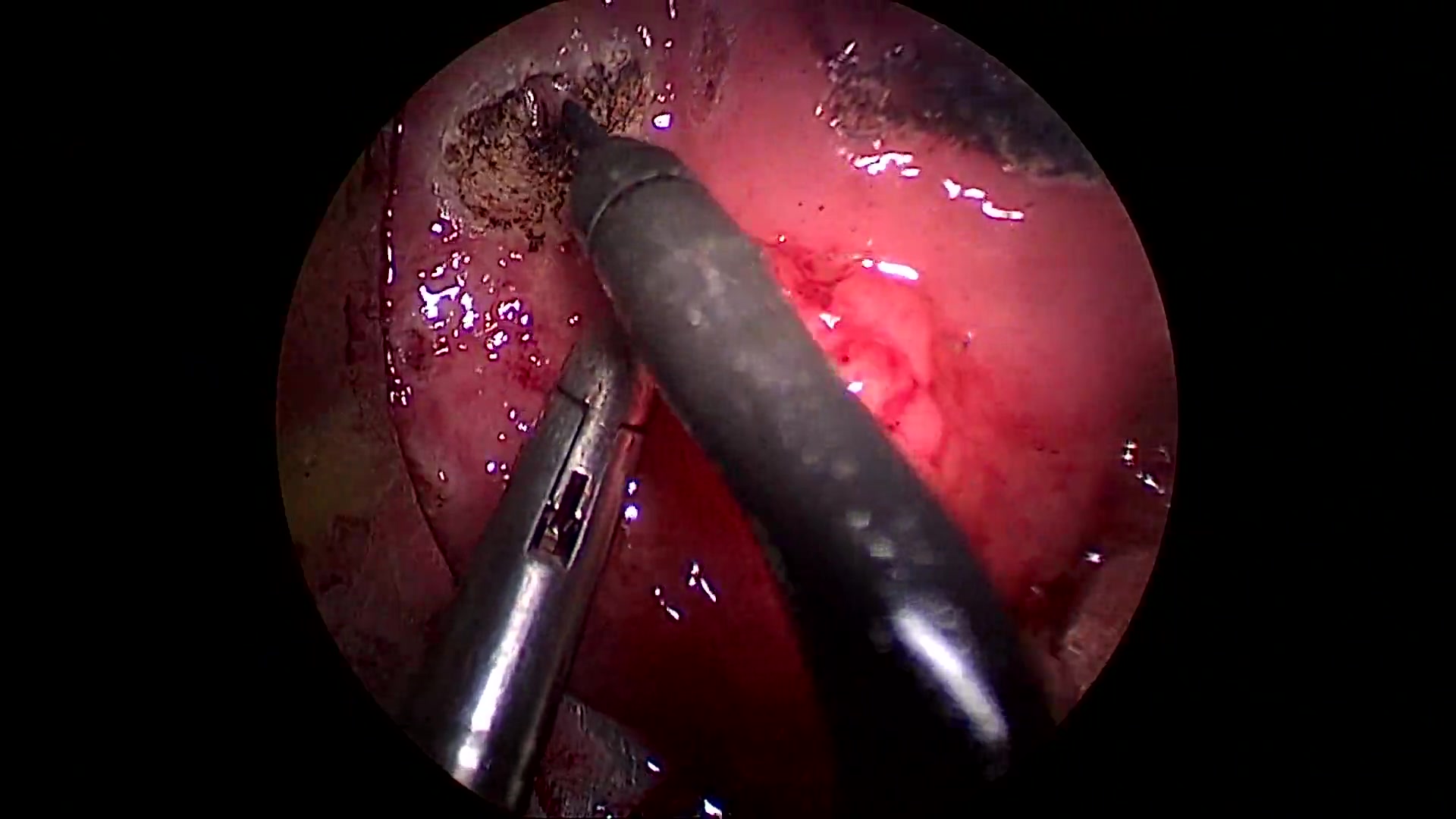} &
\sqimg{0.145\textwidth}{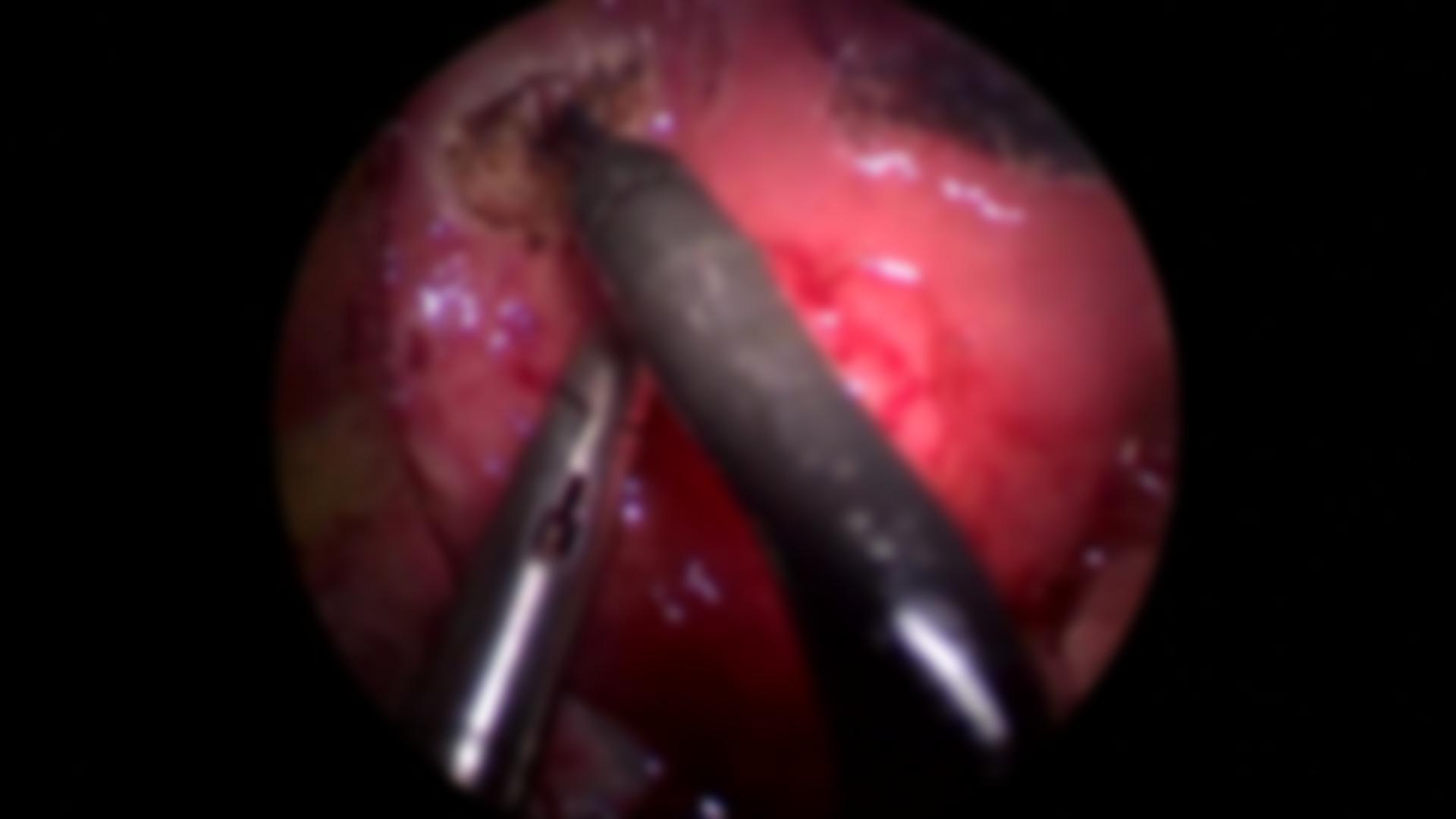} &
\sqimg{0.145\textwidth}{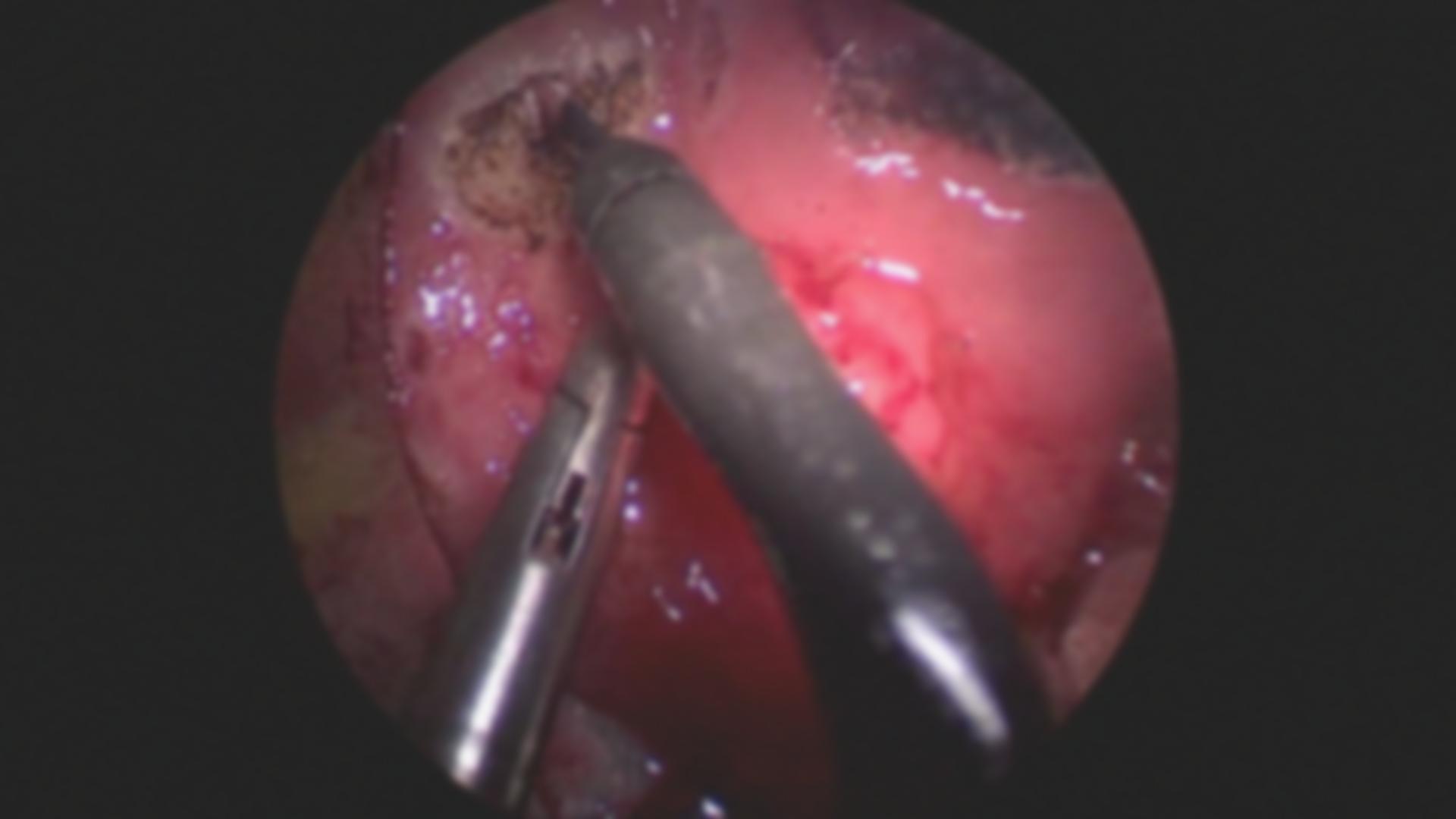} &
\sqimg{0.145\textwidth}{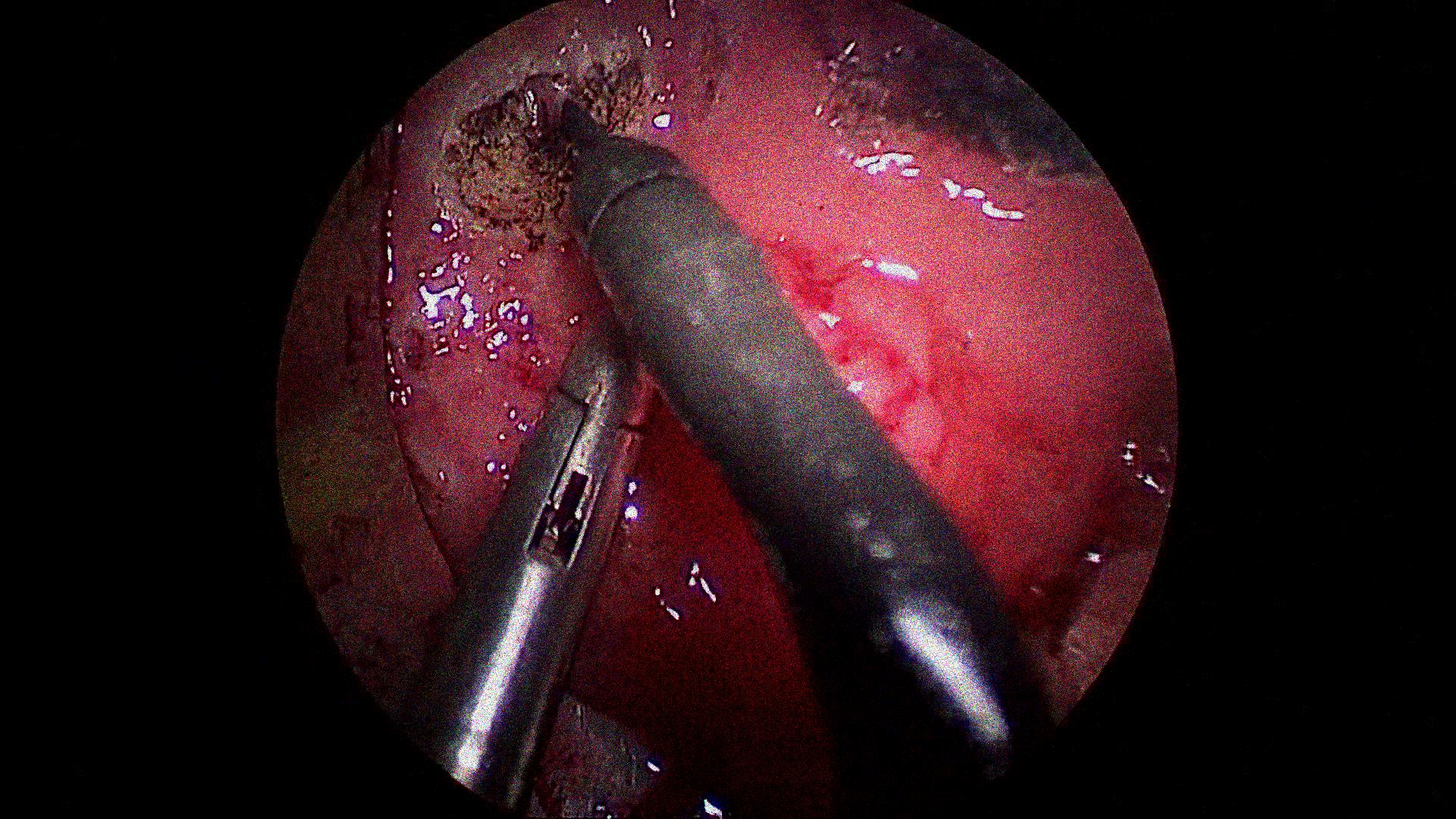} &
\sqimg{0.145\textwidth}{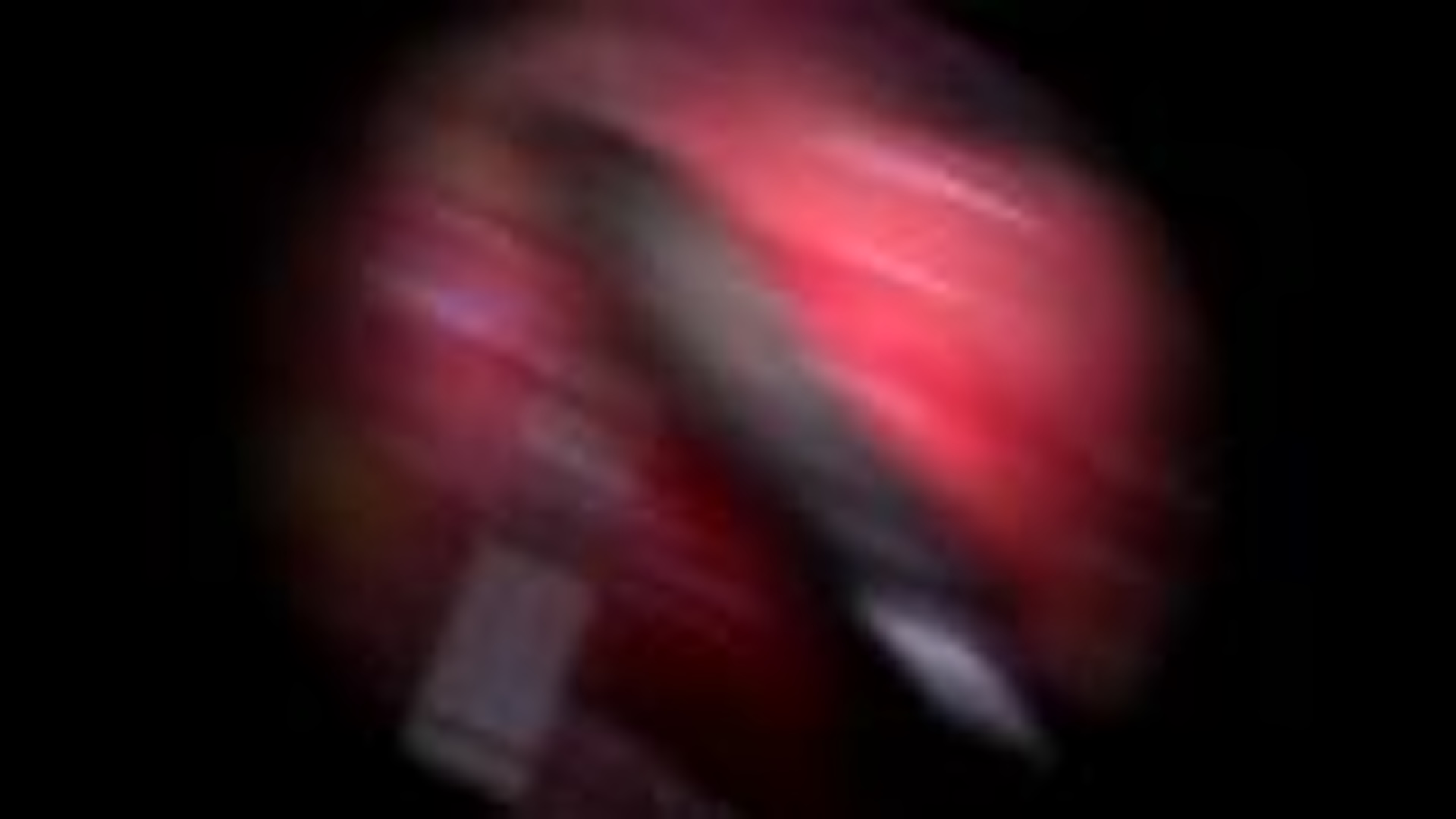} &
\sqimg{0.145\textwidth}{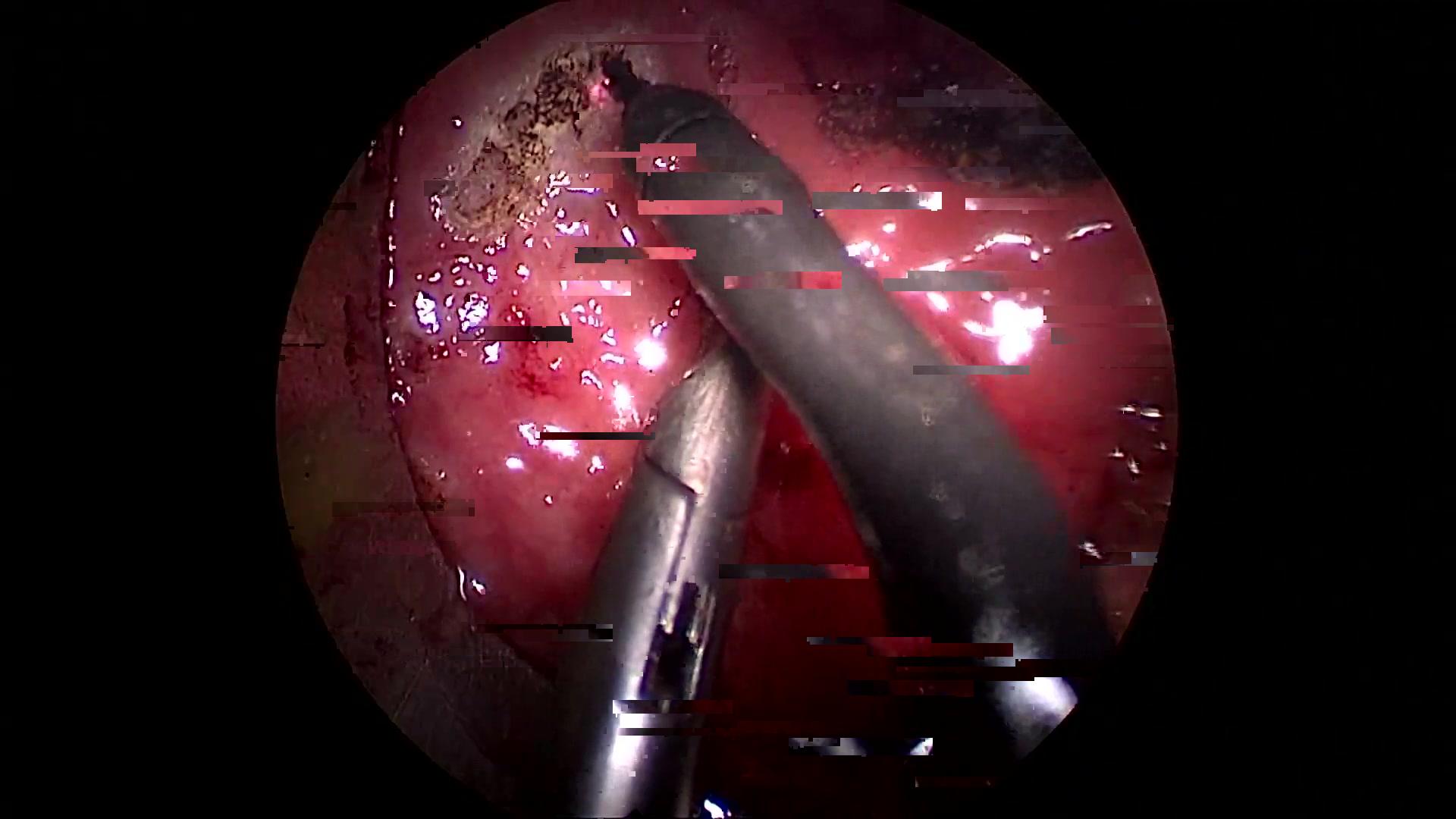} &
\sqimg{0.145\textwidth}{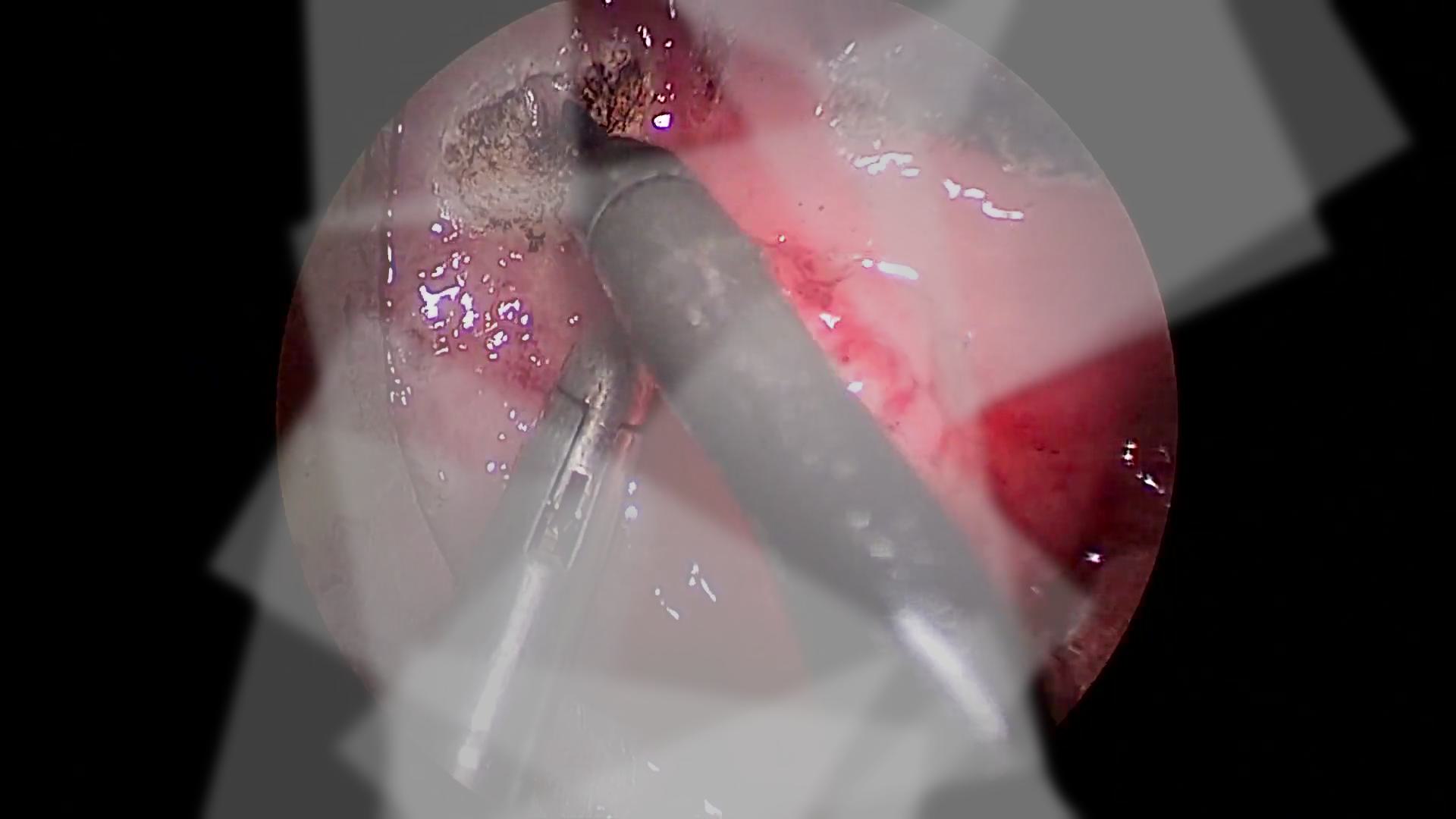} \\
% -------------------- Column labels --------------------
& \scriptsize Clean & \scriptsize Defocus & \scriptsize Fog & \scriptsize Shot & \scriptsize Motion & \scriptsize Packet & \scriptsize Smoke \\
\end{tabular}}

\caption{\textbf{Endo-C6 corrupted examples.} Clean and corrupted frames (severity 5) for the six Endo-C6 corruptions on \textit{endoscopy clips} of 3 datasets used in our benchmark.}
\label{fig:endoc6_examples_frames}

\end{figure*}

\section{Methodology}

\subsection{Endo-C6: Robustness Benchmark for Endoscopy Clips}

% A corruption benchmark is only useful if it is both representative and reproducible.
% Endoscopy meets both requirements: a small set of artifact modes recurs across centers because it is induced by the acquisition pipeline itself (optics $\rightarrow$ tissue medium $\rightarrow$ sensor $\rightarrow$ compression/streaming).
% \textbf{Endo-C6} is designed as a compact, high-stress robustness probe for temporal VLMs, focusing on six artifacts that are (i) frequent in practice, (ii) semantically label-preserving, and (iii) implementable with deterministic operators for reproducible evaluation.

\textbf{A. Benchmark taxonomy:}
Endo-C6 follows a two-axis taxonomy capturing \emph{where} artifacts arise and \emph{how} they evolve in time (Fig. \ref{fig:endoC6_taxonomy_stats}(a)).
\textbf{Domain salience} separates GI-skewed artifacts (e.g., haze/fluids), laparoscopy-skewed artifacts (e.g., cautery smoke), and domain-common effects (e.g., blur, compression/streaming).
\textbf{Temporal structure} separates quasi-static degradations (defocus, shot noise, haze) from bursty processes (motion blur spikes, packet-loss bursts, smoke plumes).
This makes failure modes easier to attribute to domain conditions and to temporal dynamics.

\noindent\textbf{B. Benchmarking protocol:}
We treat clean test clips as In-Distribution (ID) and Endo-C6 variants as controlled Out-Of-Distribution (OOD) shifts.
For each test clip $x$, we evaluate $\{x\}\cup\{c(x)\}_{c\in\text{Endo-C6}}$ on \textbf{TEMSET-24K}, \textbf{Kvasir} and \textbf{CholecT50} independently.
Corruptions are used only at test time; training and few-shot tuning (for RobustEndoCLIP) use the corresponding clean train split exclusively. Statistics are shown in Fig. \ref{fig:endoC6_taxonomy_stats}(b). 
We report \textbf{294 dataset-level evaluations} (7 adaptation \textit{settings} $\times$ 3 datasets $\times$ 7 conditions [Clean+6] $\times$ 2 stride coverages [50\%, 100\%]). This corresponds to \textbf{593{,}488 clip-level evaluations} over the Endo-C6 test splits (6{,}056 test clips $\times$ 7 conditions $\times$ 7 \textit{settings} $\times$ 2 coverages). 
Here ``7 adaptation \textit{settings}'' refer to 3 baseline models + 4 possible RobustEndoCLIP's configurations as shown in Fig.~\ref{fig:50_100_analysis}(c).
% We report Clean, Mean-C (average over corruptions), and Worst-C (minimum) to summarize average robustness and tail brittleness. 

% \footnote{Please note, TEMSET-24K is not included in Endo-C6 due to licensing, but used internally in this work to validate robustness trends.}.

% \begin{figure}[t]
% \centering
% \fbox{\includegraphics[width=0.95\linewidth]{figures/protocol_placeholder.pdf}}
% \caption{\textbf{Evaluation protocol (placeholder).} Each clean test clip is evaluated as in-distribution. Endo-C6 produces six corrupted variants per clip, simulating realistic endoscopy artifacts while keeping the label fixed.}
% \label{fig:protocol}
% \end{figure}

\subsection{RobustEndoCLIP for Temporal Endoscopy Classification}
\label{sec:methods}

\textbf{A. Zero-shot inference:}
\label{sec:prompt_infer}
We follow the CLIP interface for endoscopic videos: a temporal video encoder $f_v(\cdot)$ maps a clip $x$ to an embedding, and a text encoder $f_t(\cdot)$ maps a class prompt to the same space, as shown in Fig.~\ref{fig:main}(a).
We compute normalized embeddings
$
z_v(x)=\frac{f_v(x)}{\|f_v(x)\|},\;
z_y=\frac{f_t(t_y)}{\|f_t(t_y)\|},
$
and predict with cosine similarity and temperature $\tau$, where $\mathcal{Y}$ indexes the class set:

\begin{equation}
p(y\mid x)=
\frac{\exp(\tau\, z_v(x)^\top z_y)}{\sum_{y'\in\mathcal{Y}}\exp(\tau\, z_v(x)^\top z_{y'})}.
\label{eq:zs}
\end{equation}
All baselines (SurgVLP, HecVLP, PeskaVLP), including RobustEndoCLIP, are inferred in this manner.
We use a single canonical template $t_y=\pi(y)$ that inserts the class name into a fixed phrase (e.g., ``an endoscopy video frame of \{cls\}'').

\begin{figure}[t!]
    \centering
    \includegraphics[width=0.90\linewidth]{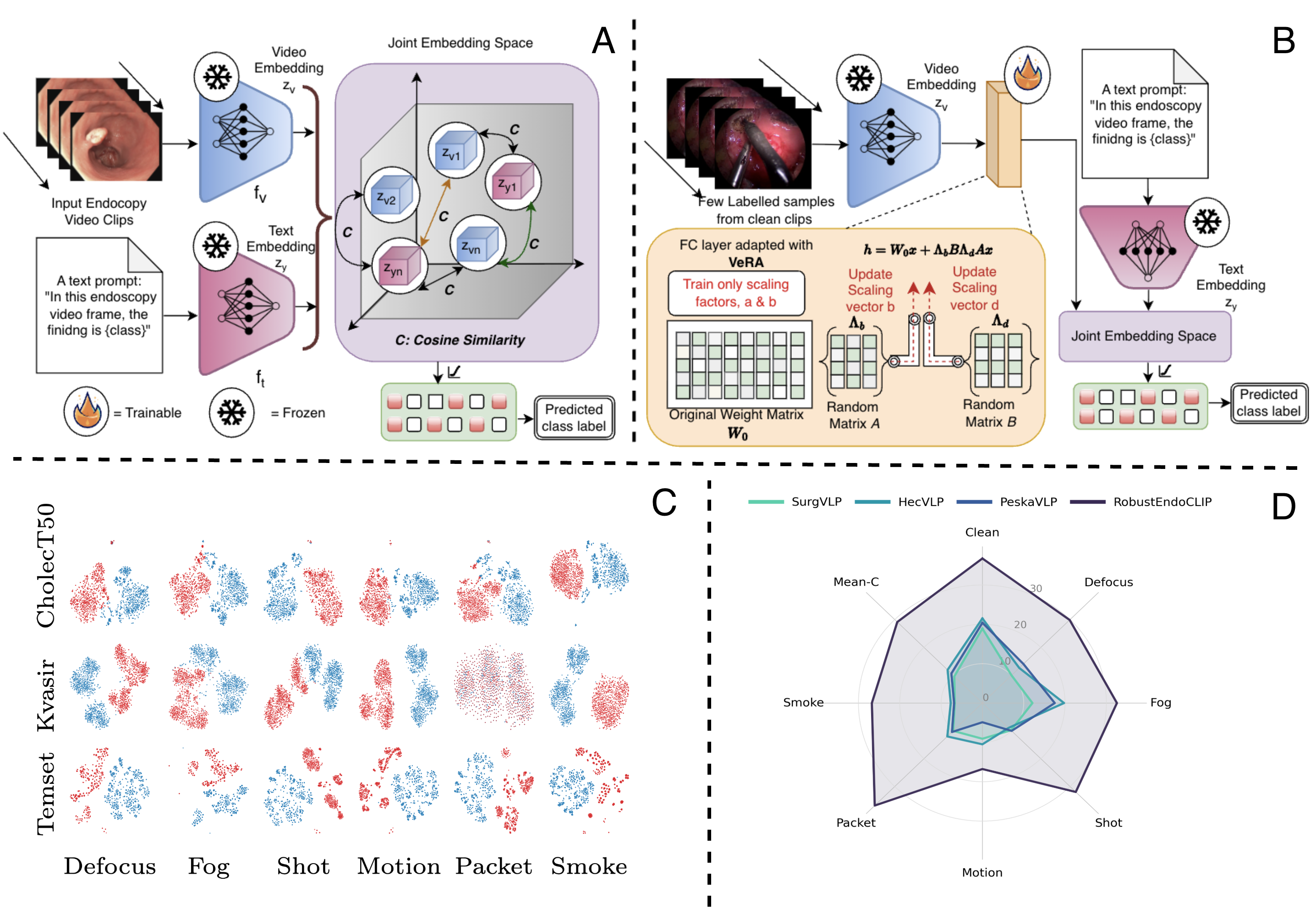} 
   
   % \caption {                  }
    \caption{
    % \textbf{Overview of RobustEndoCLIP.}
    \textbf{A)} Zero-shot inference maps an endoscopy image/frame and text class prompts into a shared embedding space. Prediction is the class whose prompt has the highest cosine similarity. \textbf{B)} Few-shot adaptation uses a small clean labeled subset from three datasets: CholecT50 (surgical phase labels), Kvasir (GI landmark/pathology class labels) and Temset24k (surgical action labels). The FC projection layer is adapted with Vector-based Random Matrix Adaptation (VeRA). Please also refer to Sec.\ref{sec:endoclip}.
    \textbf{C)} t-SNE projections of video embeddings showing how each Endo-C6 \textcolor{azure}{corruption} shifts the feature space relative to \textcolor{alizarin}{clean} clips. \textbf{D)} On average across three corruption datasets of Endo-C6, RobustEndoCLIP outperforms other TVLM baselines by a large margin. 
    }
    \label{fig:main}
   
\end{figure}

\noindent\textbf{B. RobustEndoCLIP -- Few-shot adaptation with VeRA:}
\label{sec:endoclip}
RobustEndoCLIP adapts a pretrained temporal CLIP to endoscopy with a small labeled subset of \emph{clean} training clips.
The motivation is pragmatic: zero-shot TVLMs often mis-calibrate to endoscopic appearance and label semantics, so light supervision can ``anchor'' the embedding space without retraining the backbone.

\textbf{Few-shot head tuning:} Given a labeled endoscopy dataset $\mathcal{D}\supset\{(x_i,y_i)\}_{i=1}^{N}$, we fine-tune on a limited annotated subset using a pretrained TVLM backbone.
We initialize from SurgVLP \cite{yuan2025learning} (ResNet-50 visual encoder), freeze the encoders, and update only the final FC layer using a cross-entropy objective over prompt-induced class scores:
\begin{equation}
\mathcal{L}_{\text{cls}}
=
-\sum_{(x_i,y_i)\in \mathcal{D}_s}
\log
\frac{\exp(\tau \, z_v(x_i)^\top z_{y_i})}{\sum_{y'\in\mathcal{Y}} \exp(\tau \, z_v(x_i)^\top z_{y'})}.
\label{eq:fewshot}
\end{equation}
This preserves the prompt-driven deployment interface while improving alignment to the target domain. Overall workflow is described in Fig.~\ref{fig:main}.

\textbf{Vector-based Random Matrix Adaptation (VeRA):}
LoRA adapts a weight matrix $W_0\in\mathbb{R}^{m\times n}$ using a trainable low-rank update $\Delta W=BA$ \cite{hu2022lora}.
VeRA instead fixes random matrices $A,B$ and learns only scaling vectors (Fig.~\ref{fig:main}(b)), yielding (following \cite{kopiczko2023vera})
\begin{equation}
h = W_0x + \Delta W x
= W_0x + \Lambda_b\, B\, \Lambda_d\, A\, x,
\label{eq:vera}
\end{equation}
where $\Lambda_b$ and $\Lambda_d$ are diagonal matrices formed from trainable vectors $b$ and $d$.
Intuitively, VeRA constrains the update direction and reduces the number of free parameters, which is desirable in low-shot settings to limit overfitting and stabilize tuning.
We apply VeRA to the final FC layer of the vision encoder only while keeping the backbone frozen. VeRA (being 704$\times$ lighter) directly outperforms LoRA under matched capacity as discussed in Tab.~\ref{tab:peft_ablation} and Fig.~\ref{fig:50_100_analysis}.
% \begin{algorithm}[t]
% \caption{RobustEndoCLIP: few-shot VeRA tuning on clean clips}
% \label{alg:endoclip}
% \begin{algorithmic}[1]
% \Require Pretrained TVLM $(f_v,f_t)$; class set $\mathcal{Y}$; prompt template $\pi$.
% \Require Labeled clean subset $\mathcal{D}_s$ (few-shot; e.g., 16\% of train).
% \Require Adapted layer set $\mathcal{L}$; steps $S$; learning rate $\eta$; temperature $\tau$ (fixed or learnable).
% \State Insert VeRA modules into layers $\mathcal{L}$ (freeze base weights); initialize $(b,d)$.
% \For{$s=1$ to $S$}
%     \State Sample minibatch $\{(x_i,y_i)\}$ from $\mathcal{D}_s$
%     \State Compute video embeddings $z_v(x_i)$ and prompt embeddings $\{z_y\}_{y\in\mathcal{Y}}$
%     \State Compute $\mathcal{L}_{\text{cls}}$ via Eq.~\eqref{eq:fewshot}
%     \State Update $(b,d)$ (and optionally $\tau$) by gradient descent
% \EndFor
% \State \textbf{Output:} RobustEndoCLIP, evaluated with the same prompt scoring in Eq.~\eqref{eq:zs}
% \end{algorithmic}
% \end{algorithm}

% \textbf{Training constraint:}
% To preserve the ID/OOD interpretation of Endo-C6, we tune RobustEndoCLIP only on clean training clips.
% All corrupted variants are reserved for testing, ensuring that robustness results reflect distribution shift rather than corruption-specific fitting.

\begin{table*}[t]
\centering
\caption{\textbf{Robustness Analysis of Temporal VLMs on Endo-C6.} Accuracy (\%) on clean and corrupted test clips across three datasets. \textbf{Bold} denotes Best, while \underline{Underline} denotes Second-best.}
\label{tab:robust_all_three}
\setlength{\tabcolsep}{4pt}
\renewcommand{\arraystretch}{1.2}

% -------------------- CholecT50 --------------------
\resizebox{0.80\textwidth}{!}{
\begin{tabular}{lc|cccccc|cc|c}
\toprule
\multicolumn{11}{c}{{(a) CholecT50 / Laparoscopy (downstream)}}\\
\midrule
Temporal MVLM $\downarrow$ &
Clean &
Defocus &
Fog &
Shot &
Motion &
Packet &
Smoke &
Mean-C &
Worst-C &
$\boldsymbol{\Delta}$ \\
\midrule
SurgVLP & 40.21 & 16.15 & 21.86 & 11.39 & 7.38 & 11.58 & 7.08 & 12.57 & 7.08 & 0.00 \\
HecVLP  & \underline{44.41} & 18.96 & 32.54 & \underline{17.09} & \underline{12.78} & \underline{17.44} & \underline{7.30} & \underline{17.69} & \underline{7.30} & 5.12 \\
PeskaVLP & 39.23 & \underline{24.32} & \underline{37.59} & 14.89 & 7.74 & 13.30 & 6.91 & 17.46 & 6.91 & 4.89 \\
\rowcolor{teal!15}
RobustEndoCLIP & \textbf{47.06} & \textbf{42.09} & \textbf{42.92} & \textbf{42.66} & \textbf{16.83} & \textbf{46.30} & \textbf{38.91} & \textbf{38.29} & \textbf{16.83} & \textbf{25.72} \\
\end{tabular}}

% -------------------- Kvasir --------------------
\resizebox{0.80\textwidth}{!}{
\begin{tabular}{lc|cccccc|cc|c}
\toprule
\multicolumn{11}{c}{{(b) Kvasir / Gastrointestinal (downstream)}}\\
\midrule
Temporal MVLM $\downarrow$ &
Clean &
Defocus &
Fog &
Shot &
Motion &
Packet &
Smoke &
Mean-C &
Worst-C &
$\boldsymbol{\Delta}$ \\
\midrule
SurgVLP & 13.75 & 9.50 & 12.25 & \underline{12.80} & \textbf{14.75} & 14.08 & 10.91 & 12.38 & \underline{9.50} & 0.00 \\
HecVLP  & \underline{16.20} & \underline{15.91} & \underline{20.16} & 8.25 & \underline{13.75} & \underline{14.91} & \textbf{12.16} & \underline{14.19} & 8.25 & 1.81 \\
PeskaVLP & \underline{16.20} & 14.83 & 12.08 & 12.25 & 6.50 & \textbf{15.08} & 11.08 & 11.97 & 6.50 & -0.41 \\
\rowcolor{teal!15}
RobustEndoCLIP & \textbf{36.50} & \textbf{18.83} & \textbf{26.17} & \textbf{26.08} & 13.50 & \textbf{37.17} & \underline{11.42} & \textbf{22.20} & \textbf{11.42} & \textbf{9.82} \\
\end{tabular}}

% -------------------- TEMSET-24K --------------------
\resizebox{0.80\textwidth}{!}{
\begin{tabular}{lc|cccccc|cc|c}
\toprule
\multicolumn{11}{c}{{(c) TEMSET-24K (downstream)}}\\
\midrule
Temporal MVLM $\downarrow$ &
Clean &
Defocus &
Fog &
Shot &
Motion &
Packet &
Smoke &
Mean-C &
Worst-C &
$\boldsymbol{\Delta}$ \\
\midrule
SurgVLP & 2.82 & \underline{4.24} & 2.23 & \underline{5.17} & \underline{5.10} & \underline{4.35} & 2.07 & \underline{3.86} & \underline{2.07} & 0.00 \\
HecVLP  & 3.92 & 2.98 & \underline{6.59} & 1.49 & 4.92 & 3.67 & \underline{3.51} & \underline{3.86} & 1.49 & 0.00 \\
PeskaVLP & \underline{5.78} & 2.98 & 2.87 & 2.88 & 0.40 & 2.86 & 2.56 & 2.43 & 0.40 & -1.43 \\
\rowcolor{teal!15}
RobustEndoCLIP & \textbf{26.67} & \textbf{28.67}	& \textbf{28.59} & \textbf{27.14} & \textbf{19.98} & \textbf{27.05} & \textbf{29.92} & \textbf{26.89} & \textbf{19.98} & \textbf{23.03} \\
% RobustEndoCLIP & \textbf{29.48} & \textbf{14.11} & \textbf{20.62} & \textbf{30.12} & \textbf{7.17} & \textbf{27.05} & \textbf{9.92} & \textbf{18.17} & \textbf{7.17} & \textbf{14.31} \\
\bottomrule
\end{tabular}}
\vspace{-0.3cm}
\end{table*}
% We tune on a fixed 16\% subset of the clean training split unless stated otherwise, and keep all other training data unused to preserve the ID/OOD interpretation of Endo-C6.
\section{Experimental Setup}
\textbf{A. Implementation Setup:}
In our experiments, few-shot tuning uses a ResNet-50 visual backbone (SurgVLP instantiation).
We use \textit{few-shot} for clean-label budgets up to 16\% (0/4/8/16\% in Fig. ~\ref{fig:few_shot_ablation}}),
and keep all remaining training data unused to preserve the ID/OOD interpretation
of Endo-C6. This 16\% budget is applied per dataset (TEMSET-24K, Kvasir, and CholecT50) using each dataset’s own train split. Optimization uses Adam for 50 epochs with learning rate $10^{-3}$, batch size 16, and $224\times224$ resolution. Endo-C6 corruptions (severity 5) are applied exclusively at test time to maintain a clean-to-corrupted distribution shift.

We report Top-1 accuracy on clean clips, Mean-C (average accuracy across the six corruptions), Worst-C (lowest accuracy among the corruptions), and the robustness gap 
% $\Delta=\text{Clean}-\text{Mean-C}$, which measures the average degradation under corruption.
$\Delta=\text{Mean-C}-\text{Mean-C}_{\text{SurgVLP}}$ denotes the \emph{Mean-C gain} over SurgVLP (dataset-wise).

\noindent\textbf{B. Comparative TVLMs:}
We compare against representative temporal vision-language models used in surgical video understanding, including SurgVLP, HecVLP, and PeskaVLP. These models differ in their pretraining supervision and the surgical video-text sources used for alignment, but none are trained with explicit exposure to the Endo-C6 corruption operators. Unless explicitly specified as a tuned variant, models are evaluated in their off-the-shelf configuration, using the same clip sampling policy and the same class-name prompt template, so that observed degradation under Endo-C6 reflects distribution shifts rather than differences in evaluation protocol.

\section{Results and Discussion}
\subsection{Main Results}
\label{sec:main_results}
Tab.~\ref{tab:robust_all_three} summarizes clean and Endo-C6 performance across three datasets.

\noindent\textbf{A. Corruption-wise breakdown (Q1):}
Tab.~\ref{tab:robust_all_three} shows corruption-specific failure modes.
On CholecT50, \textbf{smoke} is the dominant worst-case for baselines (Worst-C $\approx$ 6.9-7.3\%), while RobustEndoCLIP lifts the tail to 16.83\%.
On Kvasir, the weakest corruption is model-dependent (motion for PeskaVLP at 6.50, shot for HecVLP at 8.25, and defocus for SurgVLP at 9.50), indicating non-uniform sensitivity across TVLMs.
On TEMSET-24K (downstream), baselines remain unstable (Worst-C as low as 0.40), whereas RobustEndoCLIP improves accuracy across all six corruptions and raises Worst-C to 19.98\%.
Overall, the largest gains align with deployment-relevant artifacts, notably \textbf{smoke} in laparoscopy and particular improvements under \textbf{defocus/fog/shot} in GI.

\noindent\textbf{B. TVLMs across endoscopy domains (Q1, Q2):}
Failure modes differ across domains.
CholecT50 is most sensitive to \textbf{smoke} (baseline Worst-C $\approx$ 7\%), consistent with cautery-driven visibility loss, whereas GI datasets (Kvasir, TEMSET-24K) exhibit more heterogeneous worst cases spanning \textbf{motion}, \textbf{shot noise}, and \textbf{defocus} depending on the model.
RobustEndoCLIP shifts performance upward in both domains; however, the limiting corruption differs by dataset (smoke for CholecT50, motion for Kvasir, and motion for TEMSET-24K), reinforcing that endoscopy robustness is driven by domain- and dataset-specific artifact profiles rather than a single universal failure mode.

\noindent\textbf{C. Robustness of TVLMs (Q3, Q4):}
Across all three datasets, off-the-shelf TVLMs exhibit a pronounced robustness gap: corrupted accuracy and especially Worst-C can collapse even when clean accuracy is moderate.
Among baselines, HecVLP is typically strongest on Mean-C (CholecT50/Kvasir), while SurgVLP and PeskaVLP show sharper worst-case failures depending on the dataset.
RobustEndoCLIP achieves the highest Mean-C and Worst-C across datasets while preserving the prompt-based interface, indicating that lightweight few-shot adaptation can improve both average robustness and the tail.

\begin{figure}[t]
    \centering
    \includegraphics[width=0.9\linewidth]{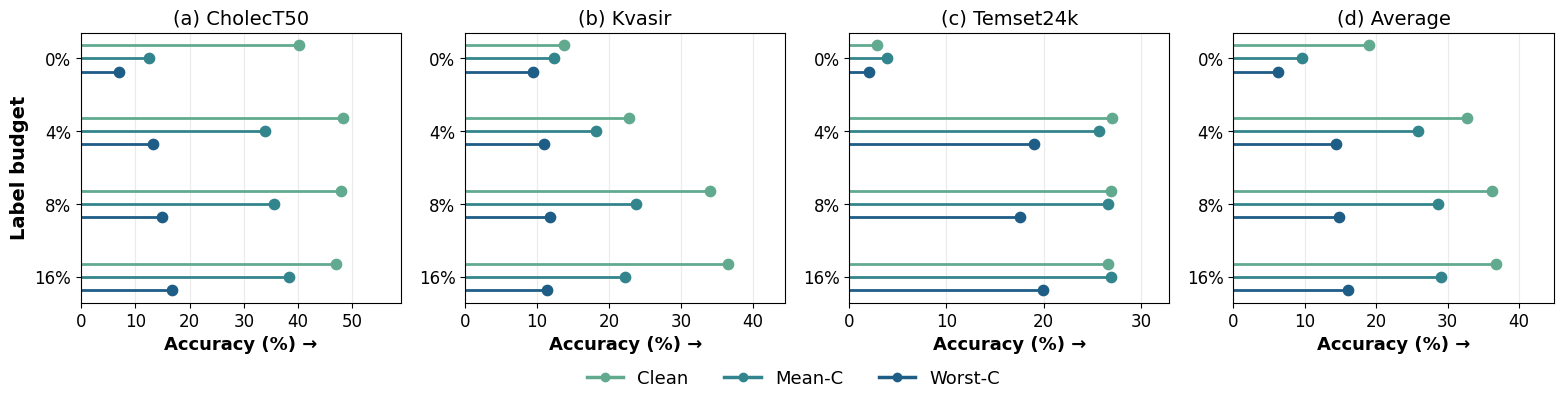}
    \vspace{-0.2cm}
    \caption{\small\textbf{Label budget scaling.} We compare the robustness of RobustEndoCLIP on different Endo-C6 datasets when tuning with different labeled fractions of the \emph{clean} training set. We report Clean, Mean-C, and Worst-C across datasets.}
    \label{fig:few_shot_ablation}
    \vspace{-0.3cm}
\end{figure}

\begin{table}[t]
\centering
\caption{\small\textbf{PEFT ablation.} Comparison of LoRA vs VeRA (and zero-shot) under a fixed label budget, reported across datasets. For each dataset we report Clean and Mean-C, and we also report the Average across datasets. \textbf{Best}. \underline{Second-best}.}
\label{tab:peft_ablation}
\setlength{\tabcolsep}{4pt}
\renewcommand{\arraystretch}{1.2}
\resizebox{0.8\textwidth}{!}{
\begin{tabular}{lcc|cc|cc|cc|cc}
\toprule
{Adaptation} & {\# Params} & {Label Budget} &
\multicolumn{2}{c}{{CholecT50}} &
\multicolumn{2}{c}{{Kvasir}} &
\multicolumn{2}{c}{{TEMSET-24K}} &
\multicolumn{2}{c}{{\textbf{Average}}} \\
\cmidrule(lr){4-5}\cmidrule(lr){6-7}\cmidrule(lr){8-9}\cmidrule(lr){10-11}
& & &
{Clean} & {Mean} &
{Clean} & {Mean} &
{Clean} & {Mean} &
{Clean} & {Mean} \\
\midrule
Zero-shot & 0 & 0 &
40.21 & 12.57 &
13.75 & 12.38 &
2.82 & 3.86 &
18.93 & 9.60 \\
LoRA & 720896 & 16\% & 
\textbf{54.18} & \underline{18.31} &
\underline{15.92} & \underline{14.29} &
\textbf{29.48} & \underline{18.17} &
\underline{33.19} & \underline{16.92} \\
% \rowcolor{teal!15}
\textbf{VeRA} & \textbf{1024} & 16\% &
\underline{47.06} & \textbf{38.29} &
\textbf{36.50} & \textbf{22.20} &
\underline{26.67} & \textbf{26.89} &
\cellcolor{teal!15}\textbf{36.74} & \cellcolor{teal!15}\textbf{29.13} \\
\bottomrule
\end{tabular}}
\vspace{-0.3cm}
\end{table}

\subsection{Discussion and Analyses}
\label{sec:analysis}

\textbf{Effect of adaptation (VeRA vs.\ LoRA) (Q4):}
Tab.~\ref{tab:peft_ablation} compares zero-shot, LoRA, and VeRA under the same 16\% label budget with head-only adaptation.
VeRA delivers the strongest robustness, achieving the best Mean-C on each dataset and the best average Mean-C (29.13).
LoRA improves clean accuracy (notably on CholecT50) but yields smaller Mean-C gains, indicating weaker robustness transfer under Endo-C6.

\textbf{Label budget scaling (0/4/8/16\%) (Q4):}
Fig.~\ref{fig:few_shot_ablation} shows that supervision improves robustness, but with dataset-dependent returns.
On CholecT50, Mean-C and Worst-C increase steadily with label budget.
On Kvasir, clean accuracy continues to rise, while Mean-C saturates earlier.
On TEMSET-24K, few-shot tuning quickly lifts both clean and corrupted accuracy, reducing the clean-corrupted gap.

\begin{figure}[t]
    \centering
    \includegraphics[width=0.9\linewidth]{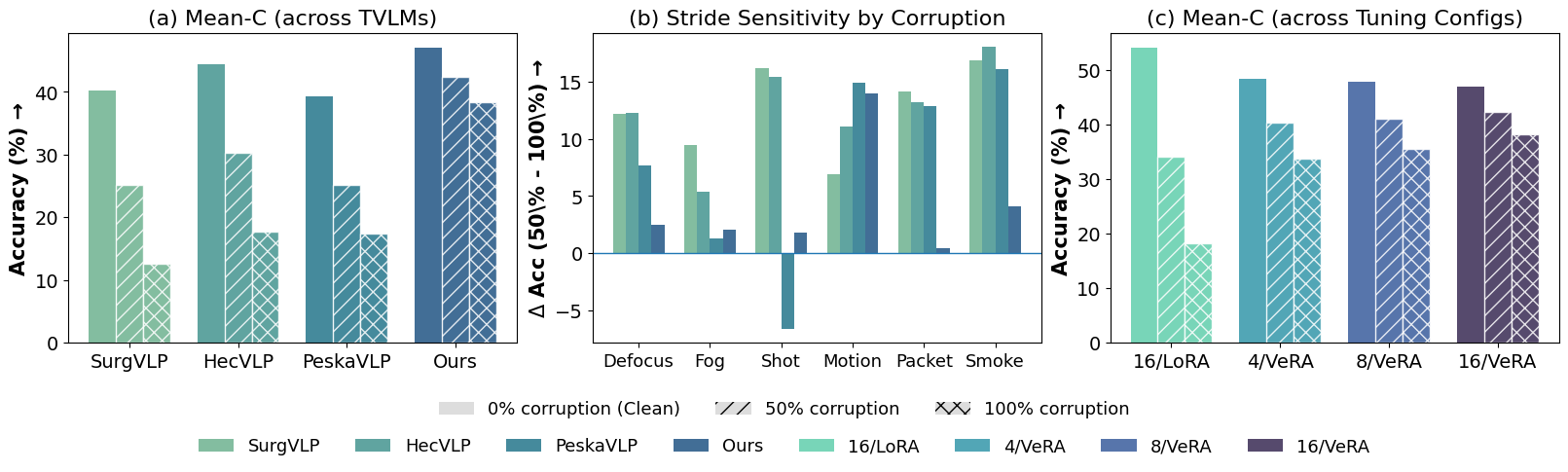}
    \vspace{-0.2cm}
    \caption{\small
        \textbf{Corruption stride analysis.}
        (a) Mean-C for temporal VLMs under clean clips and random corruption applied to 50\% vs.\ 100\% of clip frames. 
        (b) Per-corruption stride sensitivity, $\Delta\mathrm{Acc}=\mathrm{Acc}(50\%)-\mathrm{Acc}(100\%)$, across TVLMs (lower is better).
        (c) Mean-C across tuning configurations (LoRA/VeRA, varying label budgets) under different stride settings. ``Ours'' and 16/VeRA are same representing RobustEndoCLIP.
    }
    \label{fig:50_100_analysis}
    \vspace{-0.3cm}
\end{figure}

\textbf{Effect of corruption stride in clips (50\% vs.\ 100\%) (Q1, Q3):}
Fig.~\ref{fig:50_100_analysis} shows that increasing corruption coverage from 50\% to 100\% consistently lowers Mean-C.
Stride sensitivity $\Delta\mathrm{Acc}=\mathrm{Acc}(50\%)-\mathrm{Acc}(100\%)$ is largest for temporally structured artifacts (motion, packet loss, smoke).
Across tuning settings, RobustEndoCLIP's VeRA retains higher Mean-C at both coverages, indicating gains persist under denser corruption.

\section{Conclusion}
We introduced Endo-C6, a compact, fixed-severity corruption benchmark and evaluation protocol for assessing the robustness of temporal CLIP-style VLMs on endoscopy videos. Across GI and laparoscopic domains, Endo-C6 exhibits robust mean and worst-case fragility in off-the-shelf surgical TVLM baselines, while RobustEndoCLIP, a lightweight VeRA-based few-shot adaptation, substantially improves corrupted performance under matched supervision. We expect Endo-C6 to serve as a practical reporting template for robustness claims in endoscopic video understanding, emphasizing both average behavior and tail failures.

\textbf{Limitations and future work.}
This study focuses on prompt-based \emph{classification} with CLIP-style temporal VLMs; extending the benchmark to additional endoscopy tasks (e.g., action triplets, detection/segmentation) and to \emph{generative} multimodal models is a natural next step to explore.

\subsubsection*{Disclosure of Interests.} The authors have no competing interests to declare that are relevant to the content of this article.

% \bibliographystyle{splncs04}
% \bibliography{refs}

\end{document}